\documentclass[conference]{IEEEtran}
\IEEEoverridecommandlockouts
\usepackage{cite}
\usepackage{amsmath,amssymb,amsfonts}
\usepackage{graphicx}
\usepackage{textcomp}
\usepackage{xcolor}

\usepackage{algpseudocode}
\usepackage{hyperref}
\usepackage{float}
\usepackage{booktabs}
\def\BibTeX{{\rm B\kern-.05em{\sc i\kern-.025em b}\kern-.08em
    T\kern-.1667em\lower.7ex\hbox{E}\kern-.125emX}}
\begin{document}

\title{UnSegMedGAT: Unsupervised Medical Image Segmentation using Graph Attention Networks Clustering\\
}

\author{
\IEEEauthorblockN{A. Mudit Adityaja \IEEEauthorrefmark{1} ,Saurabh J. Shigwan\IEEEauthorrefmark{2} and Nitin Kumar\IEEEauthorrefmark{2}}
\IEEEauthorblockA{\IEEEauthorrefmark{1}\textit{Shiv Nadar School, Noida, India}}
\IEEEauthorblockA{\IEEEauthorrefmark{2}
\textit{Shiv Nadar Institution of Eminence, Delhi NCR, India}\\
 mudit.adityaja@sns.edu.in,\{saurabh.shigwan, nitin.kumar\}@snu.edu.in}
}

\maketitle

\begin{abstract}
The data-intensive nature of supervised classification drives the interest of the researchers towards unsupervised approaches, especially for problems such as medical image segmentation, where labeled data is scarce. Building on the recent advancements of Vision transformers (ViT) in computer vision, we propose an unsupervised segmentation framework using a pre-trained Dino-ViT~\cite{caron2021emerging}. In the proposed method, we leverage the inherent graph structure within the image to realize a significant performance gain for segmentation in medical images. For this, we introduce a modularity-based loss function coupled with a Graph Attention Network (GAT) to effectively capture the inherent graph topology within the image.
Our method achieves state-of-the-art performance, even significantly surpassing or matching that of existing (semi)supervised technique such as MedSAM~\cite{ma2024segment} which is a \textit{Segment Anything Model} in medical images. We demonstrate this using two challenging medical image datasets ISIC-2018~\cite{codella2019skin} and CVC-ColonDB~\cite{bernal2012towards}. This work underscores the potential of unsupervised approaches in advancing medical image analysis in scenarios where labeled data is scarce.
The github repository of the code is available on \url{https://github.com/mudit-adityaja/UnSegMedGAT}.
\end{abstract}


\section{Introduction}

Image segmentation is a crucial downstream task in healthcare and computer vision, with many methods developed primarily using supervised deep learning techniques that require extensive annotated data. This is particularly challenging in medical image segmentation due to the difficulty of obtaining such annotated data. We propose a novel method in unsupervised image segmentation, an area that has been comparatively less explored.
Convolutional Neural Networks (CNNs) have become popular for image segmentation, with architectures like UNet~\cite{ronneberger2015u}, SegNet~\cite{badrinarayanan2017segnet}, and DeepLab~\cite{chen2017deeplab} effectively extracting features from input images. However, these supervised methods depend on labeled datasets, which are resource-intensive to create and often limited by privacy issues, particularly in healthcare. Additionally, models trained on specific datasets may not generalize well to new data due to variations in imaging protocols and subjects. The MedSAM  model~\cite{ma2024segment}, a recent advancement in medical image segmentatin, uses a vision transformer (ViT)~\cite{caron2021emerging} architecture and is trained on a large dataset to improve generalization across diverse anatomical structures and conditions.
Despite its advancements, MedSAM struggles with modality imbalance, leading to bias towards majority classes. Moreover, it has primarily been tested on cancer data only. Recent approaches combining ViT with Graph Convolution Networks (GCN)~\cite{kipf2016semi} have shown promising results~\cite{han2022vision},\cite{aflalo2023deepcut,reddy2024unsegarmanetunsupervisedimagesegmentation} in computer vision.
In our work, we utilize a Dino-ViT~\cite{caron2021emerging} model to extract pretrained features from images and create a complete graph structure from these features. We prune edges based on features dissimilarity, and optimize segmentation clusters using Modularity loss~\cite{tsitsulin2023graph} in an unsupervised manner. The contribution of this paper can be summarized below:
\begin{itemize}
\item We propose a novel unsupervised approach in medical imaging domain by incorporating  modularity\cite{newman2006modularity} and attention mechanism \cite{vaswani2017attention} based criteria. This approach leverages features extracted from a pre-trained Vision Transformer~\cite{caron2021emerging} along with the intrinsic graph structure of the images for efficient image segmentation.
\item We deploy an ensemble of multi-level Graph attention networks~\cite{velickovic2017graph} for underlying graph clustering required for effective image segmentation.
\item 
Our method, \textit{UnSegMedGAT}, demonstrates superior performance compared to other state-of-the-art techniques, including MedSAM, on the ISIC-2018 dataset. Additionally, it significantly surpasses existing unsupervised methods on the CVC-ColonDB dataset.

\end{itemize}
\begin{figure*}
    \centering    
    \includegraphics[width=\linewidth,trim={0 12 0 0}]{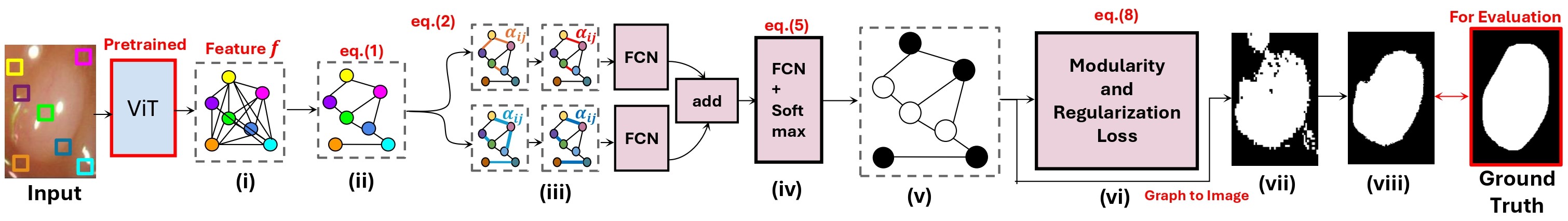}
    \caption{UnSegMedGAT Pipeline: we i) extract features $f$ of all (overlapping) image patches using vision transformer (ViT) and formulate a (complete) Graph $\mathcal{G}$ (few nodes shown, for illustration, in the same color as image patch windows), ii) then apply similarity (normalized $ff^T$) threshold to select important edges in $\mathcal{G}$, iii) aggregate and normalize features in GAT, darker node colors represent aggregation per series of GATs , iv) apply a fully connected network (FCN) with softmax activation to finally obtain node level clusters. vi) The modularity and regularization-based loss is finally used to train the model. vii-viii) At inference, edge refinement~\cite{barron2016fast} is used over the predicted mask.}
    \label{fig:method}
\end{figure*}

\section{Proposed Method}
Our approach starts with an input image of dimensions $s \times t$ and $d$ channels, which is processed through a pre-trained small vision transformer (ViT)\cite{caron2021emerging} without any fine-tuning. The transformer segments the image into patches of size $p \times p$, yielding a total of $n=st/p^2$ patches. This is followed by extracting the internal representation for each patch from the key layer features of the final transformer block, known for its strong performance across various tasks. This results in feature vectors $f$ of size $(st/p^2) \times C_{in}$, where $C_{in}$ denotes the token embedding dimension for each image patch.
Next, we construct the graph  $\mathcal{G}$ with nodes representing corresponding feature vectors $f$, that captures neighborhood relationships through a correlation matrix $A$ , defined as follows:
\begin{equation}
\label{eq:thresh_corr}
A= ff^T\cdot\biggr(ff^T > \tau\biggr) \in \mathbb{R}^{\frac{st}{p^2}\times \frac{st}{p^2}},
\end{equation}
In this equation, the entries of the adjacency matrix $A$, denoted as $A_{ij}$, take values in $(0,1)$, with $\tau \in (0,1)$ being a user-defined parameter tailored to the specific dataset. The application of equation (\ref{eq:thresh_corr}) on the normalized node features $f$ of  $\mathcal{G}$ is illustrated in Fig.~\ref{fig:method} (ii).

In case of Graph Attention Networks (GAT), we leverage an attention mechanism to dynamically weigh the importance of different nodes and their connections during the learning process. This leads to a significant improvement over traditional graph convolutional networks (GCNs), which treat all neighbors equally~\cite{kipf2016semi}.

Each node \( i \) in the graph is represented by a feature vector \( \mathbf{h}_i \in \mathbb{R}^F \), where \( F \) is the number of features.
The attention coefficients \( \alpha_{ij} \) between nodes \( i \) and \( j \) are computed using a shared attention mechanism\cite{velickovic2017graph} as follows:
\begin{equation}
\alpha_{ij} = \frac{\exp(\text{LeakyReLU}(\mathbf{a}^\top [\mathbf{W} \mathbf{h}_i || \mathbf{W} \mathbf{h}_j]))}{\sum_{k \in \mathcal{N}(i)} \exp(\text{LeakyReLU}(\mathbf{a}^\top [\mathbf{W} \mathbf{h}_i || \mathbf{W} \mathbf{h}_k]))}
\end{equation}
Here
\( \mathbf{W} \) is a learnable weight matrix,
\( \mathbf{a} \) is a learnable parameter vector,
\( || \) denotes concatenation,
\( \mathcal{N}(i) \) denotes the neighbors of node \( i \).

The output feature vector for each node \( i \) after applying the attention mechanism is computed as follows:
\begin{equation}
\mathbf{h}'_i = \sigma\left(\sum_{j \in \mathcal{N}(i)} \alpha_{ij} \mathbf{W} \mathbf{h}_j\right)
\end{equation}
Where
\( \sigma \) is a non-linear activation function (e.g., ReLU).
The sum aggregates the features of neighboring nodes weighted by their respective attention coefficients.

\subsection{Multi-Head Attention}
To enhance the model's expressiveness, GATs employ multi-head attention~\cite{vaswani2017attention}. The final output of the $i^{th}$ node can be written as concatenation of outputs from individual heads:

\begin{equation}
\mathbf{H}'_i = ||_{z=1}^{Z} \left(\sigma\left(\sum_{j \in \mathcal{N}(i)} \alpha_{ij}^z \mathbf{W}^z \mathbf{h}_j\right)\right)
\end{equation}

Here, \( Z \) represents the number of attention heads, and each head has its own weight matrix, allowing for diverse representations. \( || \) denotes concatenation operator.

\subsection{\textit{UnSegMedGAT} model}
\label{sec:model}
We employ $R$ parallel series of Graph Attention Networks (GATs) where any $r^{th}$ row consists of $t_{r}$ sequential Graph Attention layers, producing $m_{r}$ output dimensions. This allows us to capture diverse set of features at each layer. The features are subsequently aggregated through two fully connected network (FCN) layers as mentioned in equation  (\ref{eq:cluster_C}): 
(i)The first FCN projects the $m_r$ dimensions to $M$ dimensions.
(ii)The outputs from all parallel GAT layers are summed, followed by a second FCN that transforms the aggregated features from $M$ dimensions to $k$ dimensions, yielding the final cluster assignment $C \in [0,1]^{n\times k}$.
\begin{equation}
    \begin{aligned}
        \mathbf{H}'^{(r)} &= FCN(GAT^{(r)}(\hat{A},X) \\
        C &= softmax\left(FCN\left(\sum_{r=1}^R \mathbf{H}'^{(r)}\right)\right)
    \end{aligned}
    \label{eq:cluster_C}
\end{equation}
Here $n=\frac{st}{p^2}$, and $k$ are the number of nodes and clusters
respectively. 

\subsection{Loss function}
It is based on the modularity matrix $B$, which is defined on an undirected graph $\mathcal{G} = (\mathcal{V}, \mathcal{E})$, where $\mathcal{V} = (v_1, \ldots, v_n)$ represents the set of $n$ nodes and $\mathcal{E} \subseteq \mathcal{V} \times \mathcal{V}$ denotes the edges. Let $A$ be the adjacency matrix of the undirected graph $\mathcal{G}$ defined in equation (\ref{eq:thresh_corr}).

The modularity matrix $B$ is then defined as $B = A - \frac{dd^T}{2m},$
where $d$ is the degree vector of the graph $\mathcal{G}$ and $m = |\mathcal{E}|$ represents the total number of edges. The matrix $B$ quantifies the disparity between the actual edges in $\mathcal{G}$ and those expected in a random graph with an identical degree sequence. Positive entries in $B$ indicate a greater density of edges within a cluster, thus enhancing the modularity score.

The modularity $Q$ for a partition $C$ of the graph can be computed as follows:
\begin{equation}
Q = \frac{1}{2m}\sum_{ij}\left[A_{ij}-\frac{d_i d_j}{2m}\right]\delta(c_i, c_j),
\end{equation}

where $c_i$ denotes the cluster assignment for node $i$, and $\delta$ is defined by the Kronecker delta function\cite{kozen2007indefinite}.

A higher modularity score signifies a more effective partitioning of the graph $\mathcal{G}$, making the maximization of $Q$ synonymous with improved clustering. However, it has been established that maximizing this equation is an NP-hard problem~\cite{tsitsulin2023graph}. Consequently, a relaxed version of this equation expressed as:
\begin{equation}
\Bar{Q}= \frac{1}{2m}Tr(C^TBC),
\end{equation}

is commonly employed for optimization. Here, the cluster assignment matrix $C \in \mathbb{R}^{n\times k}$ (where $n$ and $k$ represent the number of nodes and clusters, respectively) is derived using equation (\ref{eq:cluster_C}). The trace operator, denoted by $Tr(\cdot)$, computes the sum of the diagonal elements of a square matrix.

We aim to cluster the resulting patch features from the ViT by minimizing the loss function $\mathcal{L}$, as delineated in~\cite{tsitsulin2023graph}:
\begin{equation}
\mathcal{L} = -\frac{1}{2m}Tr(C^TBC) + \frac{\sqrt{k}}{n}\left\lVert \sum_{i=1}^{n} C_i  \right\rVert_F - 1.
\end{equation}

In this context, $C_i$ denotes the soft cluster assignments for the $i^{th}$ node, represented as a vector of values ranging from 0 to 1 with size $k$. The notation $|| . ||_F$ refers to the Frobenius norm, which quantifies the magnitude of a matrix.

\section{Experiments and Results}
We employ the DINO-ViT small model \cite{caron2021emerging} with a patch size of $8$, and node features of size $C_{in}=384$. 

The \textit{UnSegMedGAT} model is individually optimized for each image using the loss function $\mathcal{L}$. The initial learning rate is configured to $10^{-3}$ with a decay factor of $10^{-2}$. We employ the ADAM~\cite{kingma2014adam} for the optimization of $\mathcal{L}$ in $60$ epochs. Through experimentation, we find that setting the threshold parameter $\tau = 0.3$ yields improved accuracy and faster convergence across various datasets. Additionally, we explore different activation functions such as SiLU~\cite{elfwing2018sigmoid}, SeLU~\cite{ramachandran2017searching}, and ReLU~\cite{glorot2011deep}, observing that SiLU is particularly effective for our application. 
As mentioned in subsection~\ref{sec:model}, we observed best results for $R=2$, $\{t_r\}_{r=1}^{r=R} = 2$ and $(m_1=128,m_2 = 64, M = 64)$.

We have also designed a collective approach in combination to regular optimization of UnSegMedGAT. Previously, model parameters are initialized and optimized separately per image without any sharing. In our approach, for first $\mathcal{T}_0$ epochs, all the images in a particular dataset will optimize with the same shared model parameters. After that for the remaining $\mathcal{T}_1$ epochs, model parameters are optimized separately for each image without any sharing.
Here $\mathcal{T}_0$ and $\mathcal{T}_1$ are the free parameters.
From now on, we are referring to this approach in combination to UnSegMedGAT as UnSegMedGATc.

We have considered two datasets for our experiments:1) ISIC-2018~\cite{codella2019skin} which contains dermatology images for skin cancer detection. 2) CVC-ColonDB~\cite{bernal2012towards} is a well-known open-access dataset for colonoscopy research.
For UnSegMedGAT, we found that optimal results were achieved after 300 epochs for the ISIC2018 dataset and 60 epochs for the ColonDB dataset respectively. We observed that parameter settings of $\mathcal{T}_0=5$ and $\mathcal{T}_1=5$ for ISIC2018, as well as $\mathcal{T}_0=15$ and $\mathcal{T}_1=55$ for ColonDB, yielded improved performance for UnSegMedGATc.
\begin{figure*}[ht!]
    \centering
    \includegraphics[width=0.93\linewidth,trim={0 0cm 0cm 0},clip]{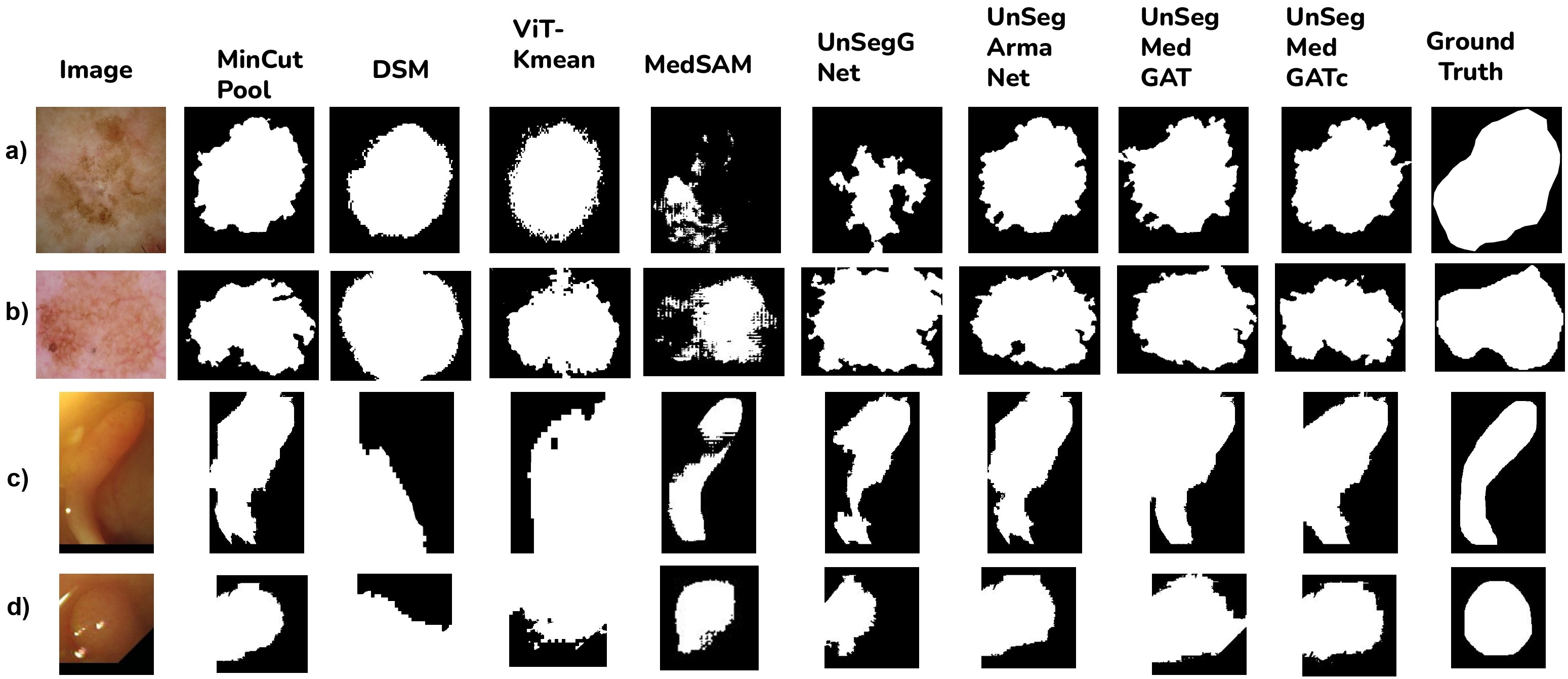}
    \caption{Segmentation results on  (a)-(b) ISIC-2018,  (c)-(d) CVC-ColonDB sample images}
    \label{fig:enter-label}
\end{figure*}
%
When comparing our UnSegMedGAT model to medical image datasets, we used the following methods:  MedSAM \cite{ma2024segment}, MinCutPool\cite{bianchi2020spectral}, GDISM\cite{trombini2023goal}, DSM\cite{melas2022deep} and ViT-Kmeans. As shown in Table I, UnSegMedGAT outperforms state-of-the-art (SOTA) unsupervised methods. Notably, UnSegMedGAT achieves significantly better scores on ISIC-2018 compared to MedSAM. However, MedSAM performs well on  CVC-ClonlonDB due to its training on large polyp datasets. On the other hand, MedSAM struggles with ISIC-2018 due to modality-imbalance issues. We also explored the impact of different activation functions on UnSegMedGAT, including SiLU, SeLU and ReLU. UnSegMedGAT with SiLU activation function yields the best results on both the medical image datasets.
We evaluate the resulting segmentation masks using the Mean Intersection over Union (mIOU) score, which is based on binarized values. A higher mIOU value indicates better segmentation accuracy, with higher values indicating better performance.
\begin{table*} [htb!]
\centering
\setlength{\tabcolsep}{2pt}
\caption{Average (mIOU, Dice) scores for medical image data}
\resizebox{\textwidth}{!}{%
\begin{tabular}{l|cccccccc}
 \textbf{Datasets} &
\textbf{UnSegMedGATc} & \textbf{UnSegMedGAT} & \textbf{UnSeGArmaNet\cite{reddy2024unsegarmanetunsupervisedimagesegmentation}} & \textbf{UnSegGNet \cite{reddy2024unseggnet}}& \textbf{MedSAM}~\cite{ma2024segment} & \textbf{MinCutPool~\cite{bianchi2020spectral}} & \textbf{DSM}~\cite{melas2022deep} & \textbf{ViT-Kmeans} \\ 
\midrule \hline
ISIC-2018 & (74.76,85.17) & (73.75,84.34) & (73.16,83.16) & (73.94,84.12) & (61.36,73.06) & (72.31,83.24) & (72.2,81.71) & (68.6,78.37) \\  
ColonDB & (65.1,77.31) & (57.21,70.76) & (59.35,72.58) & (55.67,69.18) & (70.29,80.50) & (56.09,69.71) & (47.46,63.50) & (51.84,67.23)  \\
\end{tabular}
}
\label{table:tab1}
\end{table*}
As illustrated in Table~\ref{table:tab1}, UnSegMedGAT demonstrates superior performance compared to other unsupervised methods. It achieves results on par with MedSAM for the CVC-ColonDB~\cite{bernal2012towards} and ISIC-2018~\cite{codella2019skin} datasets, despite MedSAM being fine-tuned on an extensive array of medical image datasets, including sizable colonoscopy datasets. However, MedSAM encounters challenges related to modality imbalance, which may explain its enhanced performance on the CVC-ColonDB dataset while resulting in significantly lower scores on the ISIC-2018 dataset. This imbalance likely arises from MedSAM's training on a larger polyp images in conjunction with smaller dermoscopy image-mask pairs.

\section{Conclusions}
We proposed an unsupervised image segmentation procedure capable of combining the benefits of the attention-based features captured by a ViT, and attention coefficients capturing the inherent graph structure of the images using a Graph Attention Network. Further, we also deployed a novel collective approach to update the  parameters of underlying Graph Attention network.
Our experiments on the benchmark datasets show state-of-the-art results, both on ISIC-2018 and CVC-ColonDB datasets for image segmentation. The proposed model is lightweight and hence, is less prone to overfitting problems. Further, the proposed unsupervised model provides a useful way to solve various tasks in biomedical image analysis, where getting annotated data is difficult. In future work, we intend to improve the feature representation capabilities of Graph Attention Networks at the multi-hop level by incorporating generalized modularity criteria into the loss function. This strategy is anticipated to substantially enhance segmentation accuracy across medical image datasets from diverse modalities. 


\bibliographystyle{IEEEtran}
\bibliography{reference}
\end{document}